%% file: arxiv.tex
\documentclass[10pt,twocolumn,letterpaper]{article}

\usepackage{iccv}
\usepackage{times}
\usepackage{epsfig}
\usepackage{graphicx}
\usepackage{amsmath}
\usepackage{amssymb}
\usepackage{float}
\usepackage{subfigure}
\usepackage{multirow}
\usepackage{booktabs}
\usepackage{placeins}
\usepackage{color, colortbl}
\newcommand\best[1]{\textcolor{red}{\textbf{#1}}}
\newcommand\second[1]{\textcolor{blue}{\textbf{#1}}}

\usepackage[pagebackref=true,breaklinks=true,letterpaper=true,colorlinks,bookmarks=false]{hyperref}

\iccvfinalcopy 

\def\iccvPaperID{3356} 

\ificcvfinal\fi

\def\p{{\mathbf p}}
\def\h{{\mathbf h}}
\def\z{{\mathbf z}}
\def\F{{\mathbf F}}
\def\W{{\mathbf W}}
\def\o{{\mathbf o}}
\def\MG{{\mathcal G}}

\begin{document}

\title{RangeDet: In Defense of Range View for LiDAR-based 3D Object Detection}

\author{
Lue Fan$^{1,3}$*\quad
Xuan Xiong$^{2}$*\quad
Feng Wang$^{2}$\quad
Naiyan Wang$^{2}$\quad
Zhaoxiang Zhang$^{1,3,4}$\\
$^{1}$ University of Chinese Academy of Sciences \qquad
$^{2}$ TuSimple\\
$^{3}$ Center for Research on Intelligent Perception and Computing, CASIA\\
$^{4}$ Center for Excellence in Brain Science and Intelligence Technology, CAS\\
{\tt\small
\{fanlue2019, zhaoxiang.zhang\}@ia.ac.cn \{xiongxuan08, feng.wff, winsty\}@gmail.com
}
}

\maketitle
{\let\thefootnote\relax\footnote{*The first two authors contribute equally to this work.}}
\ificcvfinal\thispagestyle{empty}\fi

\begin{abstract}
   \input{abstract.tex}
\end{abstract}

\section{Introduction}
\input{introduction_v2}
\section{Related Work}
\input{related_work}
\section{Review of Range View Representation}
\input{review_range_v2.tex}
\section{Methodology}
\input{method}
\section{Experiments}
\input{experiments}

\vspace{-2mm}
\section{Conclusion}
We present RangeDet, a range-view-based detection framework consisting of Meta-Kernel, Range Conditioned Pyramid, and weighted NMS. With our special designs, RangeDet utilizes the nature of range view to overcome a couple of challenges. RangeDet achieves comparable performance with state-of-the-art multi-view-based detectors.

{\small
\bibliographystyle{ieee_fullname}
\bibliography{egbib}
}

\end{document}

%% file: abstract.tex
In this paper, we propose an anchor-free single-stage LiDAR-based 3D object detector -- RangeDet. The most notable difference with previous works is that our method is purely based on the range view representation. 
Compared with the commonly used voxelized or Bird's Eye View (BEV) representations, the range view representation is more compact and without quantization error. Although there are works adopting it for semantic segmentation, its performance in object detection is largely behind voxelized or BEV counterparts.
We first analyze the existing range-view-based methods and find two issues overlooked by previous works: 1) the scale variation between nearby and far away objects; 2) the inconsistency between the 2D range image coordinates used in feature extraction and the 3D Cartesian coordinates used in output.
Then we deliberately design three components to address these issues in our RangeDet.
We test our RangeDet in the large-scale Waymo Open Dataset (WOD). Our best model achieves \textbf{72.9/75.9/65.8} 3D AP on vehicle/pedestrian/cyclist. These results outperform other range-view-based methods by a large margin ($\sim 20$ 3D AP in vehicle detection), and are overall comparable with the state-of-the-art multi-view-based methods. Codes will be public.

%% file: introduction_v2.tex
\begin{figure}[t]
\vspace{-0.2cm}
        \centering
         \includegraphics[height=6.5cm]{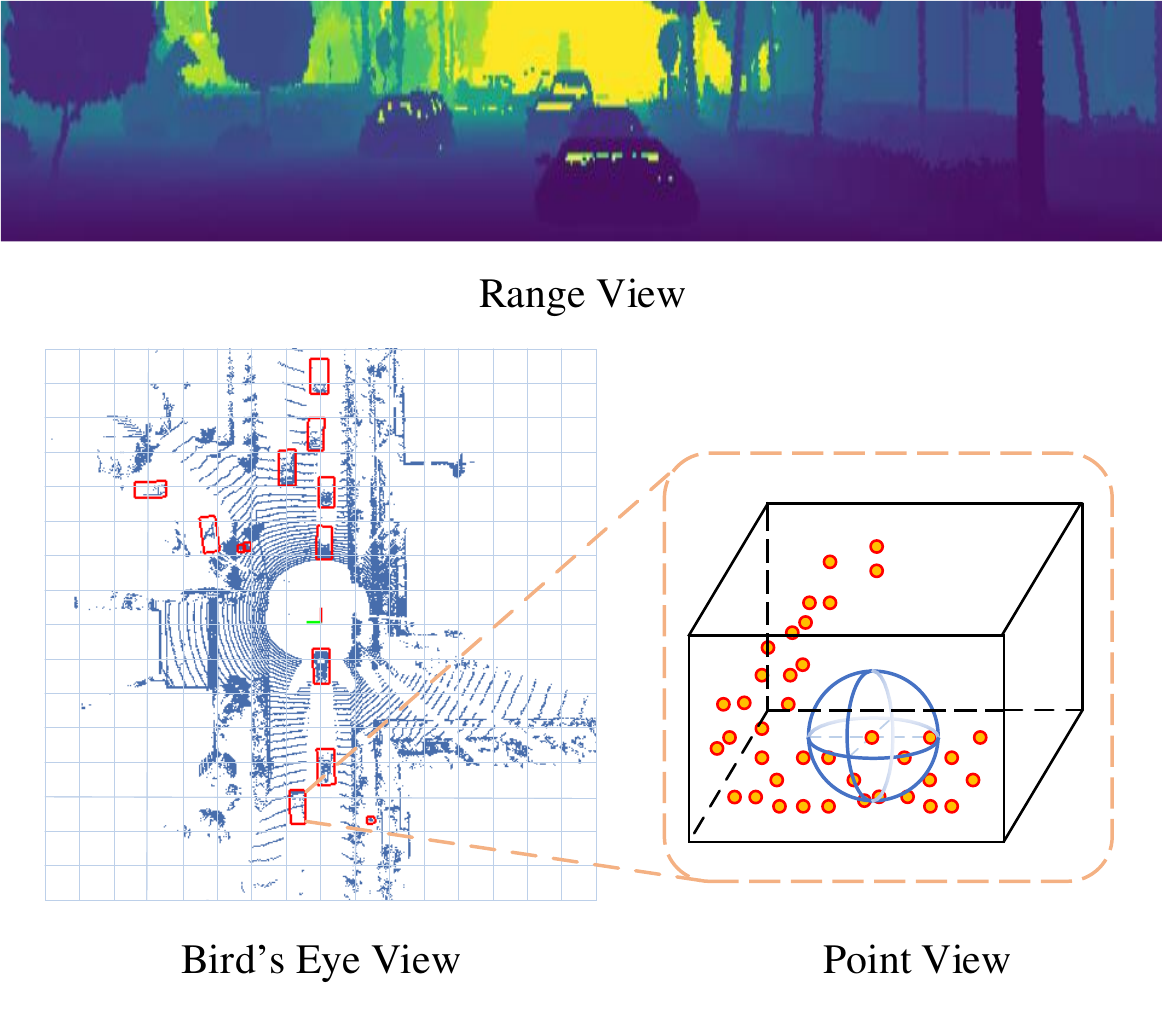}
\vspace{-0.2cm}
           \caption{Different views in LiDAR-based 3D object detection.}
           \label{fig:fig1}
\vspace{-0.2cm}
\end{figure}
LiDAR-based 3D object detection is an indispensable technology in the autonomous driving scenario. Though shared some similarities, object detection in the 3D sparse point cloud is fundamentally different from its 2D counterpart.
The key is to efficiently represent the sparse and unordered point clouds for subsequent processing.
Several popular representations include Bird's Eye View (BEV)~\cite{pointpillar,pixor,second}, Point View (PV)~\cite{pointrcnn}, Range View (RV)~\cite{velofcn,lasernet} and fusion of them~\cite{pvrcnn,mvf,pillarbased}, which are shown in Fig.\ref{fig:fig1}.
Among them, BEV is the most popular one.
However, it introduces quantization error when dividing the space into the voxels or pillars, which is unfriendly for the distant objects that may only have few points.
To overcome this drawback, the point view representation is usually incorporated. Point view operators~\cite{pointnet,pointnet++,dgcnn,cc,pointconv,kpconv,rsconv} can extract effective features from unordered point clouds, but they are difficult to scale up to large-scale point cloud data efficiently in autonomous driving scenes.
\par
The range view is widely adopted in semantic segmentation task~\cite{rangenet++, squeezesegv3, polarnet, cylinder3d}, but it is rarely used in object detection task individually.
However, in this paper, we argue that the range view itself is the most compact and informative way for representing the LiDAR point clouds because it is generated from a single viewpoint. It essentially forms a 2.5D~\cite{hu2020you} scene instead of a full 3D point cloud. Consequently, organizing the point cloud in range view misses no information. The compactness also enables fast neighborhood queries based on range image coordinates, while point view methods usually need a time-consuming ball query algorithm~\cite{pointnet++} to get the neighbors. Moreover, the valid detection range of range-view-based detectors can be as far as the sensor's availability, while we have to set a threshold for the detection range in BEV-based 3D detectors. Despite its advantages, an intriguing question raised, \emph{Why are the results of range-view-based LiDAR detection inferior to other representation forms?}
\par
Indeed some works have made attempts to make use of the range view representation from the pioneering work VeloFCN~\cite{velofcn} to LaserNet~\cite{lasernet} to the recently proposed RCD~\cite{rcd}. 
However, there is still a huge gap between the pure range-view-based method and the BEV-based method. For example, on the validation split of Waymo Open Dataset (WOD)~\cite{wod}, they are still lower than state-of-the-art methods by a large margin (more than 20 points 3D AP in vehicle class).
\par
To liberate the power of range view representation, we examine the designs of the current range-view-based detectors and found several overlooked facts. These points seem simple and obvious, but we find that the devils are in the details. Properly handling these challenges is the key to high-performance range-view-based detection.
\par
\textbf{First}, the challenge of detecting objects with sparse points in BEV is converted to the challenge of scale variation in the range image. Though there have been many methods~\cite{fpn, tridentnet} in 2D object detection tried to address this issue, this challenge is never seriously considered in the range-view-based 3D detector.
\par
\textbf{Second}, unlike in 2D image, though the convolution on range image is conducted on 2D pixel coordinates, while the output is in the 3D space.
This point suggests an inferior design in the current range-view-based detectors: both the kernel weight and aggregation strategy of standard convolution ignore this inconsistency, which leads to severe geometric information loss even from the very beginning of the network.
\par
\textbf{Third}, the 2D range view is naturally more compact than 3D space, which makes feature extractions in range-image-based detectors more efficient. However, how to utilize such characteristics to improve the performance of detectors is ignored by current range-image-based designs.
\par
In this paper, we propose a pure range-view-based framework -- RangeDet, 
which is a single-stage anchor-free detector designated to address the aforementioned challenges. 
We analyze the defects of the existing range-view-based 3D detector and point out the aforementioned three key challenges that need to be addressed.
For the first challenge, we propose a simple yet effective \emph{Range Conditioned Pyramid} to mitigate it.
For the second challenge, we propose \emph{Meta-Kernel} to capture 3D geometric information from 2D range view representation.
For the third one, we use \emph{weighted Non-Maximum Suppression} to remedy the issue.
In addition to these techniques, we also explore how to transfer common data augmentation techniques from 3D space to the range view.
Combining all the techniques, our best model achieves comparable results with state-of-the-art works in multiple views.
And we surpass previous pure range-view-based detectors by a margin of 20 3D AP in vehicle detection.
Interestingly, in contrast to common belief, RangeDet is more advantageous for farther or small objects than BEV representation.

%% file: related_work.tex
\noindent
{\bf BEV-based 3D detectors.} 
Several approaches for LiDAR-based 3D detection discretize the whole 3D space.
3DFCN~\cite{3dfcn} and PIXOR~\cite{pixor} encode handcrafted features into voxels, while VoxelNet~\cite{voxelnet} is the first to use end-to-end learned voxel features. SECOND~\cite{second} accelerates VoxelNet by sparse convolution. PointPillars~\cite{pointpillar} is aggressive in feature reduction that it applies PointNet to collapse the height dimension first and then treat it as a pseudo-image.

\noindent
{\bf Point-view-based 3D detectors.} 
F-PointNet~\cite{frustumpointnet} first generates frustums corresponding to 2D Region of Interest (ROI), then use PointNet~\cite{pointnet} to segment foreground points and regress the 3D bounding boxes. 
PointRCNN~\cite{pointrcnn} generates 3D proposals directly from the whole point clouds instead of 2D
images for 3D detection with point clouds by using PointNet++~\cite{pointnet++} both in proposal generation and refinement. IPOD~\cite{ipod} and STD~\cite{std} are both two-stage methods which use the foreground point cloud as a seed to generate proposals and refine them in the second stage.

\noindent
{\bf Range-view-based 3D detectors.} VeloFCN~\cite{velofcn} is a pioneering work in range image detection, which projects point cloud to 2D and applies 2D convolutions to predict 3D box for each foreground point densely. 
LaserNet~\cite{lasernet} uses a fully convolutional network to predict a multimodal distribution for each point to generate the final prediction. Recently, RCD~\cite{rcd} addresses the challenges in range-view based detection by learning a dynamic dilation for scale variation and soft range gating for the ``boundary blur'' issue as pointed in Pseudo-LiDAR~\cite{pseudolidar}. 

\noindent
{\bf Multi-view-based 3D detectors.}
MV3D~\cite{mv3d} is the first work to fuse features in frontal view, BEV, and camera view for 3D object detection.
PV-RCNN~\cite{pvrcnn} jointly encodes point and voxel information to generate high-quality 3D proposals.
MVF~\cite{mvf} endows a wealth of contextual information from different perspectives for each point to improve the detection of small objects.

\noindent
{\bf 2D detectors.}
Scale variation is a long-standing problem in 2D object detection. SNIP~\cite{snip} and SNIPER~\cite{sniper} rescale proposals to a normalized size explicitly based on the idea of image pyramids. FPN~\cite{fpn} and its variants~\cite{panet,librarcnn} build feature pyramids, which have become the indispensable component for modern detectors. TridentNet~\cite{tridentnet} constructs weight-shared branches but using different dilation to build scale-aware feature maps.

%% file: review_range_v2.tex
\label{sec:review}
In this section, we quickly review the range view representation of LiDAR data.
\par
\begin{figure}[h]
\vspace{-0.2cm}
	\centering
	\includegraphics[width=0.7\linewidth]{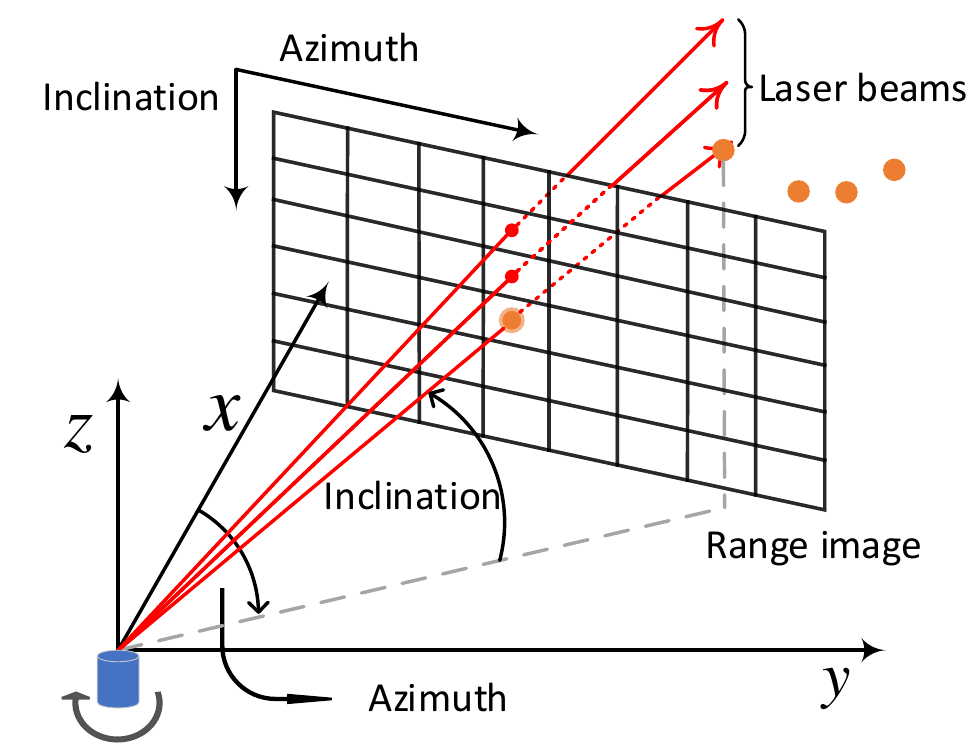}
    \caption{The illustration of the native range image.}
    \label{fig:azimuth}
\vspace{-0.2cm}
\end{figure}

For a LiDAR with $m$ beams and $n$ times measurement in one scan cycle, the returned values from one scan form a $m \times n$ matrix, called \textbf{range image} (Fig. \ref{fig:fig1}). 
Each column of the range image shares an azimuth, and each row of the range image shares an inclination. They indicate the relative vertical and horizontal angle of a returned point w.r.t the LiDAR original point.
The pixel value in the range image contains the range (depth) of the corresponding point, the magnitude of the returned laser pulse called intensity and other auxiliary information. One pixel in the range image contains at least three geometric values: range $r$, azimuth $\theta$, and inclination $\phi$. These three values then define a spherical coordinate system.
Fig. \ref{fig:azimuth} illustrates the formation of the range image and these geometric values.
The commonly-used point cloud data with Cartesian coordinates is actually decoded from the spherical coordinate system:
\vspace{-0.2cm}
\begin{equation}
    \begin{aligned}
    x &= r\cos(\phi)\cos(\theta),\\
    y &= r\cos(\phi)\sin(\theta),\\
    z &= r\sin(\phi),\\
    \end{aligned}
\vspace{-0.2cm}
\end{equation}
where $x,y,z$ denote the Cartesian coordinates of points. Note that range view is only valid for the scan from one viewpoint. It is not available for general point cloud since they may overlap for one pixel in the range image.
\par
Unlike other LiDAR datasets, WOD directly provides the native range image. Except for range and intensity values, WOD also provides another information called elongation~\cite{wod}. The elongation measures the extent to which the width of the laser pulse is elongated, which helps distinguish spurious objects.

%% file: method.tex
\begin{figure*}[t]
        \centering
         \includegraphics[width=0.9\linewidth]{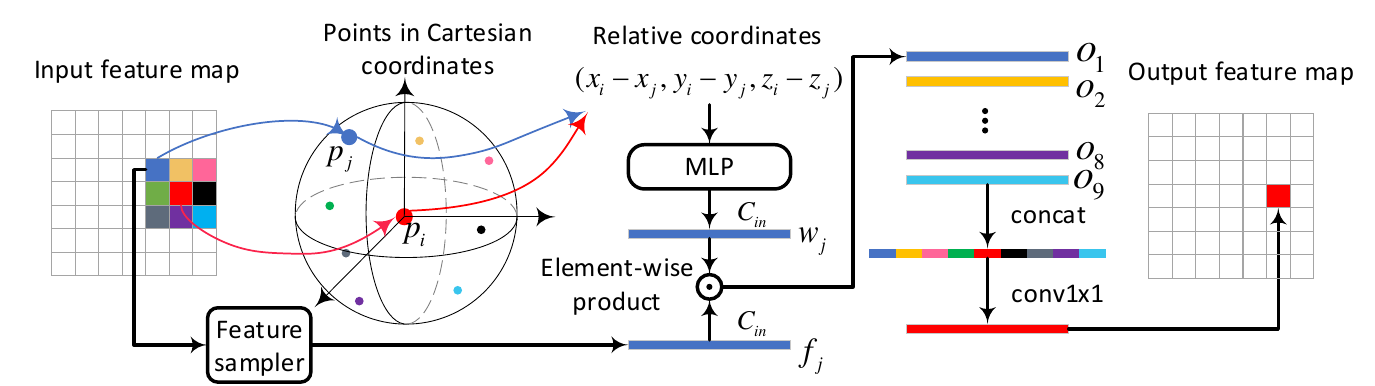}
           \caption{The illustration of Meta-Kernel (best viewed in color). Taking a 3x3 sampling grid as an example, we can get relative Cartesian coordinates of nine neighbors to the center. A shared MLP takes these relative coordinates as input to generate nine weight vectors: ${w_1, w_2, \cdots, w_9}$. Then we sample nine input feature vectors:${f_1, f_2, \cdots, f_9}$. $o_i$ is the element-wise product of $w_i$ and $f_i$. By passing a concatenation of $o_i$ from nine neighbors to a $1 \times 1$ convolution, we aggregate the information from different channels and different sampling locations and get the output feature vector.}
           \label{fig:meta}
\end{figure*}
In this section, we first elaborate on three components of RangeDet. Then the full architecture is presented.
\subsection{Range Conditioned Pyramid}\label{sec_rca}
\input{rcp.tex}
\subsection{Meta-Kernel Convolution}\label{sec_meta}
\input{meta_kernel_v4.tex}

\subsection{Weighted Non-Maximum Suppression}
\input{wnms_v2}
\label{sec_wnms}
\subsection{Data Augmentation in Range View}
\input{aug}
\subsection{Architecture}
\input{architecture}

%% file: rcp.tex
In 2D detection, feature-pyramid-based methods such as Feature Pyramid Network (FPN)~\cite{fpn} are usually adopted to address the scale variation issue. We first construct the feature pyramids as in FPN which is illustrated in Fig. \ref{fig:architecture}. 
Although the construction of the feature pyramid is similar to that of FPN in 2D object detection, the difference lies in how to assign each object to a different layer for training.
In the original FPN, the ground-truth bounding box is assigned based on its area in the 2D image.
Nevertheless, simply adopting this assignment method ignores the difference between the 2D range image and 3D Cartesian space.
A nearby passenger car may have similar area with a far away truck but their scan patterns are largely different.
Therefore, we designate the objects with a similar range to be processed by the same layer instead of purely using the area in FPN.
Thus we name our structure as Range Conditioned Pyramid (RCP). 

%% file: meta_kernel_v4.tex
\input{tables/diff_conv.tex}
Compared with the RGB image, the depth information endows range images with a Cartesian coordinate system, however standard convolution is designed for 2D images on regular pixel coordinates. For each pixel within the convolution kernel, the weights only depend on the relative pixel coordinates, which can not fully exploit the geometric information from the Cartesian coordinates. In this paper, we design a new operator which learns dynamic weights from relative Cartesian coordinates or more meta-data, making the convolution more suitable to the range image.
\par
For better understanding, we first disassemble standard convolution into four components: sampling, weight acquisition, multiplication and aggregation.

\noindent 1) \textbf{Sampling.} The sampling locations in standard convolution is a regular grid $\MG$, which has $k_h \times k_w$ relative pixel coordinates.
For example, a common $3\times 3$ sampling grid with dilation 1 is:
\vspace{-0.1cm}
\begin{equation}
    \mathcal{G} = \{(-1, -1), (-1, 0), ..., (1, 0), (1, 1)\}.
\vspace{-0.1cm}
\end{equation}
For each location $\p_{0}$ on the input feature map $\F$, we usually sample feature vectors of its neighbors $\F(\p_{0}+\p_{n})$, $\p_{n} \in \MG$ using \texttt{im2col} operation.

\noindent 2) \textbf{Weight acquisition.} The weight matrix $\W(\p_n) \in \mathbb{R}^{C_{out} \times C_{in}}$ for each sampling location $(\p_0 + \p_n)$ depends on $\p_n$, and fixed for a given feature map. This is also called the ``weight sharing'' mechanism for convolution.

\noindent 3) \textbf{Multiplication.} We decompose the matrix multiplication of the standard convolution into two steps. The first step is pixel-wise matrix multiplication. 
For each sampling point $(\p_0 + \p_n)$, its output is defined as 
\vspace{-0.0cm}
\begin{equation}
	\o_{\p_0}(\p_n) = \mathbf{W(p}_n) \cdot \F(\p_0 + \p_n).
\vspace{-0.0cm}
\end{equation}

\noindent 4) \textbf{Aggregation.} After multiplication, the second step is to sum over all the $\o_{\p_0}(\p_n)$ in $\MG$, which is called channel-wise summation. 
\par
In summary, the standard convolution can be presented as:
\begin{equation}
    \z(\p_{0}) = \sum_{\p_n \in \MG} \o_{\p_0}(\p_n).
\end{equation}
\par
In our range view convolution, we expect that the convolution operation is aware of the local 3D structure. Thus, we make the weight adaptive to the local 3D structure via a meta-learning approach.
\par
\textbf{For weight acquisition}, we first collect the meta-information of each sampling location and denote this relationship vector as $\h(\p_0, \p_n)$.
$\h(\p_0, \p_n)$ usually contains relative Cartesian coordinates, range value, etc. Then we generate the convolution weight $\W_{\p_0}(\p_n)$ based on $\h(\p_0, \p_n)$. Specifically, We apply a Multi-Layer Perceptron (MLP) with two fully-connected layers:
\begin{equation}
    \W_{\p_0}(\p_n) = \texttt{MLP}(\h(\p_0, \p_n)).
\end{equation}
\par
\textbf{For multiplication}, instead of matrix multiplication, we simply use element-wise product to obtain $\o_{\p_0}(\p_n)$ as follows: 
\begin{equation}
    \o_{\p_0}(\p_n) = \W_{\p_0}(\p_n) \odot \F(\p_0 + \p_n).
\end{equation}
We do not use matrix multiplication because our algorithm runs on large-scale point clouds, and it costs too much GPU memory to save a weight tensor with shape $H\times W \times C_{out} \times k_h \times k_w \times C_{in}$. Inspired by the depth-wise convolution, the element-wise product eliminates the $C_{out}$ dimension from the weight tensor, which is much less memory-consuming. However, there is no cross-channel fusion in the element-wise product. We leave it to the aggregation step.

\textbf{For aggregation}, instead of channel-wise summation, we concatenate all $\o_{\p_0}(\p_n)$, $\forall \p_n \in \MG$ and pass it to a fully-connected layer to aggregate the information from different channels and different sampling locations.
\par
Summing it up, the Meta-Kernel can be formulated as:
\begin{equation}
    \z(\p_0) = \mathcal{A}(\W_{\p_0}(\p_n) \odot \F(\p_0 + \p_n)),\quad \forall \p_n \in \MG,
\end{equation}
\noindent
where $\mathcal{A}$ is the aggregation operation containing concatenation and a fully-connected layer. Fig. \ref{fig:meta} provides a clear illustration of Meta-Kernel.
\par
\noindent
\textbf{Comparison with point-based operators.} Although shares some similarities with point-based convolution-like operators, Meta-Kernel has three significant differences from them.
(1) Definition space. Meta-Kernel is defined in 2D range view, while others are defined in the 3D space.
So Meta-Kernel has regular $n \times n$ neighborhood, and point-based operators have an irregular neighborhood. 
(2) Aggregation. Points in 3D space are unordered, so the aggregation step in point-based operators is usually permutation-invariant. Max-pooling and summation are widely adopted. $n \times n$ neighbors in the RV are permutation-variant, which is a natural advantage for Meta-Kernel to adopt concatenation and fully-connected layer as the aggregation step.
(3) Efficiency.
Point-based operators involve time-consuming key-point sampling and neighbor query.
For example, downsampling 160K points to 16K with Farthest Point Sampling (FPS)~\cite{pointnet++} takes 6.5 seconds in a single 2080Ti GPU, which is also analyzed in RandLA-Net~\cite{randla}.
Some point-based operators, such as PointConv~\cite{pointconv}, KPConv~\cite{kpconv} and the native version of Continuous Conv~\cite{cc}, generate a weight matrix or feature matrix for each point, so they face severe memory issue processing large-scale point cloud. 
These disadvantages make it impossible to apply point-based operators to large-scale point clouds (more than $10^5$ points) in autonomous driving scenarios. 
\par
For a clear comparison, we summarize the differences between several closely related work and our Meta-Kernel convolution in Table \ref{tab:diff_conv}.

%% file: tables/diff_conv.tex
\begin{table*}[t]
\footnotesize
\begin{center}
\begin{tabular}{l|c|c|c|c|c}
\toprule
Conv Type & Space & Sampling & Weight acquisition & Multiplication & Aggregation \\
\midrule
Standard Conv& 2D & Grid & Network parameter & Matrix multiplication & Channel-wise summation\\
\hline
Depth-wise Conv~\cite{depthwise}& 2D & Grid & Network parameter & Element-wise product & Channel-wise summation\\
\hline
RCD~\cite{rcd} & 2D &  Learned dilation & Network parameter & Matrix multiplication & Channel-wise summation\\
\hline
Deformable Conv~\cite{dcn}& 2D &  Learned offset & Network parameter & Matrix multiplication & Channel-wise summation\\
\hline
PointNet~\cite{pointnet}& 3D &  Ball query & Network parameter & Matrix multiplication & Max-pooling\\
\hline
Continuous Conv~\cite{cc}& 3D & $k$NN & Generated by MLP & Matrix multiplication & Channel-wise summation\\
\hline
EdgeConv~\cite{dgcnn}& 3D & $k$NN & Network parameter & Matrix multiplication & Max-pooling\\
\hline
RS Conv~\cite{rsconv}& 3D & Ball query & Generated by MLP & Element-wise product & Max-pooling\\
\hline
RandLA~\cite{randla}& 3D & $k$NN & Network parameter & Matrix multiplication & Attentive pooling\\
\hline
KPConv~\cite{kpconv}& 3D & Ball query & Weighted sum of parameter & Matrix multiplication & Channel-wise summation \\
\hline
Meta-Kernel & 2D &  Grid & Generated by MLP & Element-wise product & Concat and $1\times 1$ convolution\\
\bottomrule
\end{tabular}
\end{center}
\vspace{-0.2cm}
\caption{Comparison of different convolutions.}
\vspace{-0.4cm}
\label{tab:diff_conv}
\end{table*}

%% file: wnms_v2.tex
As mentioned earlier, how to utilize the compactness of range view representation to improve the performance of range-image-based detectors is an important topic.
{\color{black} In common object detectors, a proposal inevitably has a random deviation from the mean of the proposal distribution. The straightforward way to get a proposal with small deviation is to choose the one with the highest confidence. While a better and more robust way to eliminate the deviation is using the majority votes of all the available proposals. An off-the-shelf technique just fits our need -- weighted NMS~\cite{wnms}. Here comes an advantage of our method: the nature of compactness makes RangeDet feasible to generate proposals in the full-resolution feature map without huge computation cost, however it is infeasible for most BEV-based or point-view-based methods. With more proposals, the deviation will be better eliminated.}
\par
We first filter out the proposals whose scores are less than a predefined threshold 0.5, and then sort the proposals as in standard NMS by their predicted scores. For the current top-rank proposal $\mathbf{b_0}$, we find the proposals whose IoUs with $\mathbf{b_0}$ are higher than 0.5. The output bounding box for $\mathbf{b_0}$ is a weighted average of these proposals, which can be described as:
\vspace{-0.2cm}
\begin{equation}
   \mathbf{\widehat{b}_0} = \frac{\sum_k \mathbb{I}(\text{IoU}(\mathbf{b_0}, \mathbf{b_k}) > t) s_k \mathbf{b_k}}{\sum_k \mathbb{I}(\text{IoU}(\mathbf{b_0}, \mathbf{b_k}) > t) s_k},
\vspace{-0.2cm}
\end{equation}
where $\mathbf{b_k}$ and $s_k$ denote other proposals and corresponding scores. $t$ is the IoU threshold, which is 0.5. $\mathbb{I}(\cdot)$ is the indicator function. 

%% file: aug.tex
Data augmentation is a critical technique to improve the performance of LiDAR-based 3D object detectors.
\emph{Random global rotation}, \emph{Random global flip} and \emph{Copy-Paste} are three typical ones.
Although they are straightforward in 3D space, it's non-trivial to transfer them to RV and preserve the structure of RV.
\par
Rotation of point clouds can be regarded as translation of range images along the azimuth direction.
Flipping in 3D space corresponds to the flipping with respect to one or two vertical axes of range images (We provide a clear illustration in supplementary materials).
From the leftmost column to the rightmost, the span of azimuth is $(-\pi, \pi)$. So, unlike the augmentation of 2D RGB-image, we calculate the new coordinate of each point to keep it consistent with its azimuth.
For Copy-Paste~\cite{second}, the objects are pasted on the new range image with their original vertical pixel coordinates. Because of the non-uniform vertical angular resolution, we can only keep the structure of RV and avoid objects largely deviating from the ground by this treatment.
Besides, a car in the distance should not be pasted in the front of a nearby wall. So we carry out ``range test'' to avoid such a situation.

%% file: architecture.tex
\begin{figure*}[t]
        \centering
         \includegraphics[width=0.9\linewidth]{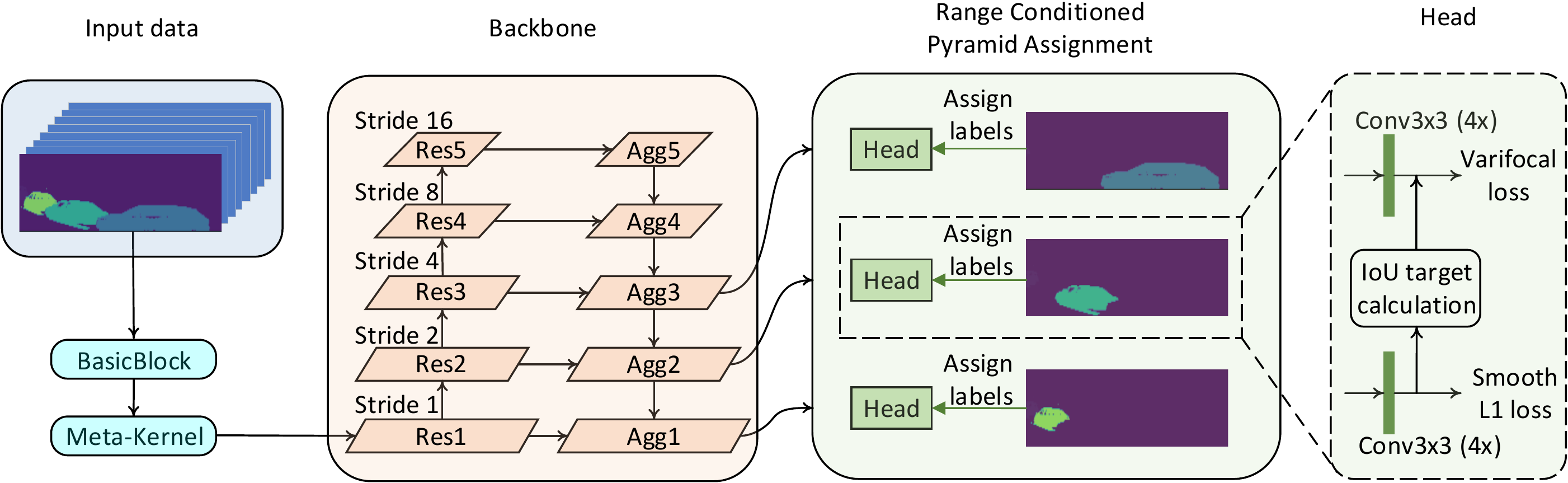}
           \caption{The overall architecture of RangeDet.}
           \label{fig:architecture}
           \vspace{-0.2cm}
\end{figure*}
\noindent
\textbf{Overall pipeline.} The architecture of RangeDet is shown in Fig. \ref{fig:architecture}. The eight input range image channels include range, intensity, elongation, x, y, z, azimuth, and inclination, as described in Sec. \ref{sec:review}. Meta-Kernel is placed in the second BasicBlock\cite{resnet}. Feature maps are downsampled to stride 16, and upsampled to full resolution gradually. Next, we assign each ground-truth bounding box to the layers of stride $1, 2, 4$ in RCP according to the range of the box center. 
All the positions whose corresponding points are in ground-truth 3D bounding boxes are treated as positive samples, otherwise negative. 
At last, we adopt Weighted NMS to de-duplicate the proposals and generate high-quality results.
\par
\noindent
\textbf{RCP and Meta-Kernel.} In WOD, the range of a point varies from 0m to 80m. According to the distribution of points in the ground-truth bounding boxes, we divide $[0, 80]$ to 3 intervals: $[0,15), [15,30), [30,80]$.
We a use two-layer MLP with 64 filters to generate weights from relative Cartesian coordinates. ReLU is adopted as activation.
\par
\input{loss.tex}

%% file: loss.tex
\noindent
\textbf{IoU Prediction head.} 
In the classification branch, we adopt a very recent work -- varifocal loss\cite{vfl} to predict IoU between the predicted bounding box and the ground-truth bounding box. Our classification loss is defined as:
\vspace{-0.2cm}
\begin{equation}
  L_{\mathrm{cls}} = \frac{1}{M} \sum_i{ \text{VFL}_{i}},
\vspace{-0.2cm}
  \end{equation}
  where $M$ is the number of valid points, and $i$ is the point index. $\text{VFL}_i$ is the varifocal loss of each point:
\vspace{-0.2cm}
\begin{equation} \label{fl}
     \begin{split}
        \text{VFL}(p,q) = \left\{
                \begin{array}{ll}
                  -q(q \log(p) + (1-q)\log (1-p)), q > 0 \\
                  -\alpha p^\gamma \log (1-p), \qquad \qquad\qquad \quad q = 0,
                \end{array}
              \right.
     \end{split}
\vspace{-0.2cm}
\end{equation}
where $p$ is the predicted score, and $q$ is the IoU between the predicted bounding box and the ground-truth bounding box. $\alpha$ and $\gamma$ play a similar role as in focal loss~\cite{focalloss}. 
\par
\begin{figure}[h]
        \centering
           \includegraphics[width=0.9\linewidth]{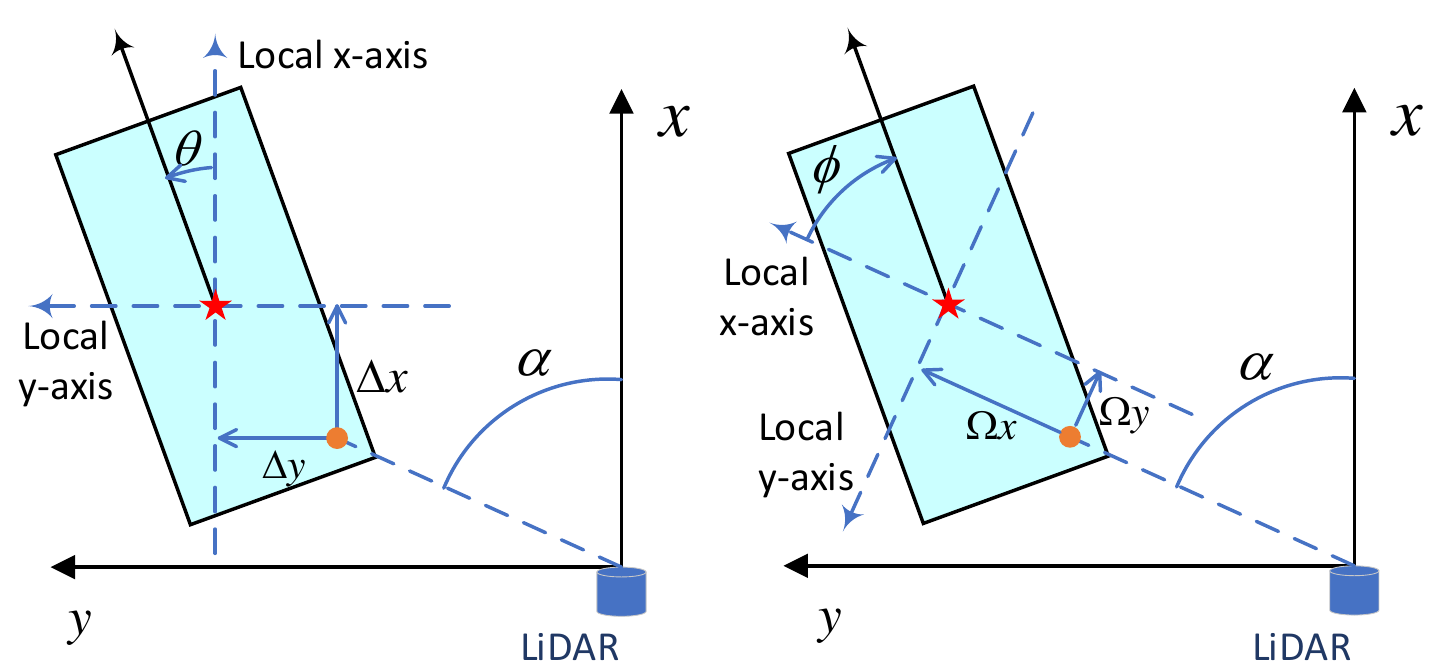}
           \caption{The illustration of two kinds of regression targets. Left: For all the points, the $x$-axis of the egocentric coordinate system is regarded as the local $x$-axis. Right: For each point, its azimuth direction is regarded as local $x$-axis. Before calculating the regression loss, we first transform the first kind of targets to the latter.}
           \label{fig:target}
\vspace{-0.2cm}
\end{figure}

\noindent
\textbf{Regression head.} 
The regression branch also contains four $3\times 3$ Conv as in the classification branch. We first formulate the ground-truth bounding box containing point\footnote{Here, a point is actually a location in the feature map and corresponds to a Cartesian coordinate. For a better understanding, we still call it a point.} $i$ as
$(x_i^g, y_i^g, z_i^g, l_i^g, w_i^g, h_i^g, \theta_i^g)$
to denote the coordinates of the bounding box center, dimension and orientation, respectively. The Cartesian coordinate of point $i$ is $(x_i, y_i, z_i)$. We define the offsets between the point $i$ and the center of bounding box containing point $i$ as
$\Delta r_i = r_i^g - r_i,\;r \in \{x, y, z\}$.
For point $i$, we regard its azimuth direction as its local $x$-axis which is the same as in LaserNet~\cite{lasernet}. And we formulate such transformation as follows (Fig. \ref{fig:target} provides a clear illustration):
\vspace{-0.2cm}
\begin{equation} \label{eqn:box-center}
\begin{aligned}
\alpha_i &=  \text{arctan2}\left(y_i, x_i \right),\\
\boldsymbol{R_i} &= 
 \left[ \begin{array}{ccc}
    \cos\alpha_i & \sin\alpha_i & 0\\
    -\sin\alpha_i & \cos\alpha_i & 0\\
    0 & 0 & 1\\
  \end{array}
  \right],\\
\phi_i^g = \theta_i^g - \alpha_i,\;\;&\left[\Omega x_i, \Omega y_i, \Omega z_i \right] =  \boldsymbol{R_i} \left[\Delta x_i, \Delta y_i, \Delta z_i \right]^\top,\\
\end{aligned}
\vspace{-0.0cm}
\end{equation}
where $\alpha_i$ denotes the azimuth of point $i$, and $\left[\Omega x_i, \Omega y_i, \Omega z_i \right]$ is the transformed coordinate offset to be regressed.
Such a transformed target is appropriate for range-image-based detection since an object's appearance in the range image doesn't change with the azimuth in a fixed range.
Thus, it's reasonable to make regression targets azimuth-invariant. So for each point, we regard azimuth direction as local $x$-axis.
\par
We denote the point $i$'s ground-truth targets set $\mathcal{Q}_i$ as $\left\{\Omega x_i^g, \Omega y_i^g, \Omega z_i^g, \log l_i^g, \log w_i^g, \log h_i^g, \cos\phi_i^g, \sin\phi_i^g \right\}$. So the regression loss is defined as
\vspace{-0.2cm}
\begin{equation}\label{eq:loss}
      L_{\mathrm{reg}}  = \frac{1}{N} \sum_i \left(  \frac{1}{n_{i}} \sum_{q_i \in \mathcal{Q}_i}\mathrm{SmoothL1}(q_i - p_i) \right),
\vspace{-0.2cm}
\end{equation}
where $p_i$ is the predicted counterpart of $q_i$. $N$ is the number of ground-truth bounding boxes, and $n_i$ is the number of points in the bounding box which contains the point $i$.
The total loss is the sum of $L_{cls}$ and $L_{reg}$.

%% file: experiments.tex
We conduct our experiments on large-scale Waymo Open Dataset (WOD), which is the only dataset that provides native range images.
We report LEVEL\_1 average precision in all experiments for comparing with other methods.
Please refer to supplemental material for the detailed results and configuration of the pipeline.
Experiments in Table \ref{tab:ablation}, Table \ref{tab:sota} and Table \ref{tab:wnms_on_diff_methods} using the entire training dataset. And we uniformly sample 25\% training data ($\sim 40$k frames) for other experiments.
\input{tables/ablation_table_wo_aug}
\vspace{-1mm}
\subsection{Study of Meta-Kernel Convolution}
\input{meta_ablation}

\vspace{-1mm}
\subsection{Study of Range Conditioned Pyramid}
\input{rcp_ablation}
\vspace{-1mm}
\subsection{Study of Weighted Non-Maximum Suppression}
\input{wnms_ablation}
\vspace{-1mm}
\subsection{Ablation Experiments}
\input{ablation}
\vspace{-1mm}
\subsection{Comparison with State-of-the-Art Methods}
\input{sota}
\vspace{-1mm}
\subsection{Runtime Evaluation}
\vspace{-1mm}
\input{runtime}

%% file: tables/ablation_table_wo_aug.tex
\begin{table*}[t]
	\footnotesize 
	\begin{center}
		\scalebox{1.0}{
			\begin{tabular}{c|ccccc|cccc|cccc}
				\toprule
				\multirow{2}{*}{} &\multirow{2}{*}{\shortstack[1]{Meta-\\Kernel}} & \multirow{2}{*}{RCP} & \multirow{2}{*}{\shortstack[1]{IoU\\Prediction}} & \multirow{2}{*}{WNMS}& \multirow{2}{*}{DA}  & 			
				\multicolumn{4}{c|}{3D AP (IoU=0.7)} & \multicolumn{4}{c}{BEV AP (IoU=0.7)} \\
				&&&&&&Overall & 0 - 30 & 30 - 50 & 50 - inf & Overall & 0 - 30 & 30 - 50 & 50 - inf\\
				\midrule
				A1& &  & & &&53.39 & 73.02 & 48.79 & 28.14 & 70.45 & 86.22 & 68.45 & 51.90 \\
				A2&\checkmark &  &  &  & &56.58 & 76.11 & 54.29& 32.53 & 74.89 & 88.14 & 71.43 & 57.55\\
				A3&&\checkmark   &  &  & &58.37 & 78.66 & 50.40& 32.35 & 78.02 & 90.71 & 73.13 & 62.23\\
				A4&\checkmark & \checkmark &  &  & &61.05 & 80.11 & 54.59 & 35.95 & 80.65 & 92.12 & 78.20 & 66.58\\
				A5&\checkmark & \checkmark & \checkmark &  &&64.61 & 84.87& 61.13 & 40.87 & 82.32 & 93.17 &80.49 & 68.98 \\
				A6&\checkmark & \checkmark & \checkmark & \checkmark & &69.00 & {86.89} & {66.16} & 45.81 & 85.48 & 93.62 & 82.17 & 72.97\\
				A7&\checkmark &  & \checkmark & \checkmark &&64.35 & 82.60 & 60.11 & 39.91 & 77.33 & 89.19 & 75.69 & 61.33\\
				A8& &  & \checkmark & \checkmark & &61.08 & 81.78 & 58.07 & 36.22 & 76.20 & 88.78 & 72.31 & 58.94\\
				A9&\checkmark & \checkmark & \checkmark & \checkmark & \checkmark & 72.85 & {87.96} & {69.03} & 48.88 & 86.94 & 94.35 & 85.66 & 77.01\\
				 \bottomrule
			\end{tabular}
		}	
	\end{center}
	\vspace{-2mm}
	\caption{Ablation of our components on vehicle detection. DA stands for data augmentation.}
	\label{tab:ablation}
	\vspace{-3mm}
\end{table*}

%% file: meta_ablation.tex
We conduct extensive experiments to ablate Meta-Kernel in this section.
These experiments do not involve data augmentation.
We build our baseline by replacing Meta-Kernel with a 2D $3 \times 3$ convolution.
\par
\noindent
\textbf{Different input features.} Table \ref{tab:diff_input} shows the results of different meta information as input.
Not surprisingly, using relative pixel coordinates (E4) only brings marginal improvements compared with the baseline, demonstrating the necessity of Cartesian information (coordinates or range) in kernel weight.
\input{tables/diff_input.tex}
\input{tables/sota_table}
\par
\noindent
\textbf{Different locations to place Meta-Kernel.} We place the Meta-Kernel at stages with different strides. The results are shown in Table \ref{tab:diff_layer}, which demonstrates that Meta-Kernel is more prominent at a lower level. This result is reasonable since the low-level layers have a closer association with geometric structure, where the Meta-Kernel takes a vital role.
\input{tables/diff_layer}
\par
\noindent
\textbf{Performance on small objects.}
Boundary information is more crucial for small objects in range view, for example pedestrian, to avoid being diluted by background than large objects. Meta-Kernel enhances the boundary information by capturing local geometric features, so it is especially powerful in small objects detection. Table \ref{tab:ped_3d} shows the significant effectiveness.
\input{tables/ped_3d.tex}
\par
\noindent
\textbf{Comparison with point-based operators.}
We discussed the main differences between Meta-Kernel and point-based operators in Sec. \ref{sec_meta}.
For a fair comparison, we implement some typical point-based operators on the 2D range image with \emph{fixed $3\times3$ neighborhood} just like our Meta-Kernel.
Please refer to supplementary materials for the implementation details.
Some operators such as KPConv~\cite{kpconv}, PointConv~\cite{pointconv} are not implemented due to huge memory costs. These methods all obtain inferior results as Table \ref{tab:point_op} shows. We owe it to the strategies they used for aggregation in unordered point clouds, which will be elaborated next. 
\input{tables/point_op.tex}
\par
\noindent
\textbf{Different ways of aggregation.} Instead of concatenation, we try max-pooling and summation in a channel-wise manner just like other point-based operators, and Table \ref{tab:diff_agg} shows the results. Performance significantly drops when using max-pooling or summation as they treat the features from different locations equally. These results demonstrate the importance of keeping and utilizing the relative orders in range view. Note that other views cannot adopt concatenation due to the disorder of point clouds.
\par
\input{tables/diff_agg.tex}

%% file: tables/diff_input.tex
\begin{table}[ht]
\small
\begin{center}
    \resizebox{0.9\linewidth}{!}{
	\begin{tabular}{c|c|c}
	\toprule
	& Meta-data &  3D AP \\
	\midrule
	E1 &Baseline & 63.57\\
	E2 &$(x_i - x_j, y_i - y_j, z_i - z_j)$ & 67.00 \\
	E3 &$(x_j, y_j, z_j)$ & 64.05 \\
	E4 &$(u_i - u_j, v_i - v_j)$ & 63.87 \\
	E5 &$(x_i, y_i, z_i, x_j, y_j, z_j)$ & 65.33 \\
	E6 &$(r_i - r_j)$ & 67.31 \\
	E7 &$(x_i - x_j, y_i - y_j, z_i - z_j, r_i - r_j)$ & 67.37 \\
	E8 &$(x_i - x_j, y_i - y_j, z_i - z_j, u_i - u_j, v_i - v_j)$ & 67.11 \\
	\bottomrule
	\end{tabular}
	}
\end{center}
\vspace{-0.2cm}
\caption{Performance comparison of different inputs for our Meta-Kernel. In baseline experiment, Meta-Kernel is replaced by a $3\times3$ 2D convolution. $(x_i, y_i, z_i)$, $(u_i,v_i)$ and $r_i$ stand for Cartesian coordinates, pixel coordinates and range, respectively.}
\label{tab:diff_input}
\vspace{-0.2cm}
\end{table}

%% file: tables/sota_table.tex
\begin{table*}[h]
	\footnotesize 
	\begin{center}
		\scalebox{0.99}{
			\begin{tabular}{c||c|cccc|cccc}
				\toprule
				\multirow{2}{*}{Method} & \multirow{2}{*}{View} & 			
				\multicolumn{4}{c|}{3D AP on Vehicle (IoU=0.7)} & \multicolumn{4}{c}{3D AP on Pedestrian (IoU=0.5)} \\
				&&Overall & 0m - 30m & 30m - 50m & 50m - inf & Overall & 0m - 30m & 30m - 50m & 50m - inf\\
				
				\midrule
				PointPillars$\ast$~\cite{pointpillar} & BEV & 56.62 & 81.01 & 51.75 & 27.94 & 59.25 & 67.99 & 57.01 & 41.29\\
				PointPillars$\dag$~\cite{pointpillar} & BEV & 62.2 & 81.8 & 55.7 & 31.2 & 60.0 & 68.9 & 57.6 & 46.0\\
				DynVox\cite{mvf} & BEV & 59.29 & 84.9 & 56.08 & 31.07 & 60.83 & 69.76 & 58.43 & 42.06\\
				MVF~\cite{mvf} & BEV + RV & 62.93 & 86.3 & 60.2 & 36.02 & 65.33 & 72.51 & 63.35 & 50.62\\
				PillarOD~\cite{pillarbased} & BEV + CV & {69.8} & {88.53} & {66.5} & {42.93} & \second{72.51} & \second{79.34} & \second{72.14} & \second{56.77}\\
				Voxel-RCNN~\cite{voxelrcnn} & BEV & \best{75.59} & \best{92.49} & \best{74.09} & \best{53.15} & - & - & - & -\\
				PointPillars$\P$~\cite{pointpillar} & BEV & 71.56 & - & - & - & 70.61 & - & - & -\\
				PV-RCNN~\cite{pvrcnn} & BEV + PV & {70.3} & \second{91.92} & \second{69.21} & 42.17 & - & - & - & -\\
				\midrule
				LaserNet~\cite{lasernet} & RV & 52.11 & 70.94 & 52.91 & 29.62 & 63.4 & 73.47 & 61.55 & 42.69\\
				RCD (the first stage)~\cite{rcd} & RV & 55.01 & - & - & - & - & - & - & -\\
				RCD~\cite{rcd} & RV + PV & 66.39 & 86.59 & 65.64 & 40.00 & - & - & - & -\\
				\midrule
				Ours & RV & \second{72.85} & {87.96} & {69.03} & \second{48.88} & \best{75.94} & \best{82.20} & \best{75.39} & \best{65.74}\\
				\bottomrule
			\end{tabular}
		}	
	\end{center}
	\vspace{-2mm}
	\caption{Results of vehicle and pedestrian evaluated on WOD validation split. Please refer to supplementary materials for detailed results of cyclist. BEV: Bird's Eys View. RV: Range View. CV: Cylindrical View~\cite{pillarbased}. PV: Point View. 
	$\dag$: implemented by \cite{wod}.
	$\P$: implemented by MMDetection3D.
	$\ast$: implemented by \cite{mvf}. The best result and the second result are marked in \best{red} and \second{blue}, respectively.}
	\label{tab:sota}
	\vspace{-2mm}
\end{table*}

%% file: tables/diff_layer.tex
\begin{table}[H]
\small
\begin{center}
    \resizebox{0.9\linewidth}{!}{
	\begin{tabular}{l|c|c|c|c|c}
	\toprule
	Stage stride &Baseline& 1 & 2 & 4 & 8\\
	\midrule
	3D AP  & 63.57 & 67.37 & 64.79 & 63.66 & 63.80\\
	\bottomrule
	\end{tabular}
	}
\end{center}
\vspace{-0.2cm}
\caption{Performances on vehicle class when Meta-Kernel is placed in different stages of different strides.}
\label{tab:diff_layer}
\vspace{-0.1cm}
\end{table}

%% file: tables/ped_3d.tex
\begin{table}[H]
	\small 
	\begin{center}
		\resizebox{0.9\linewidth}{!}{
			\begin{tabular}{c|ccccc}
				\toprule
				\multirow{2}{*}{Method} &  			
				\multicolumn{4}{c}{3D AP on Pedestrian (IoU=0.5)}\\
				&Overall & 0 - 30 & 30 - 50 & 50 - inf\\
                \midrule
				w/o Meta-Kernel &  {69.06} & {77.86} & {67.79} & {53.94}\\
                w/ Meta-Kernel &  {74.16} & {80.86} & {73.54} & {63.21}\\
                \midrule
				Improvements &  {+5.09} & {+3.00} & {+5.75} & {+9.27}\\
				\bottomrule
			\end{tabular}
		}	
	\end{center}
  \vspace{-0.2cm}
	\caption{Ablation of Meta-Kernel on pedestrian.}
	\label{tab:ped_3d}
  \vspace{-0.3cm}
\end{table}

%% file: tables/point_op.tex
\begin{table}[H]
	\small 
	\begin{center}
		\resizebox{0.98\linewidth}{!}{
			\begin{tabular}{c|ccccc}
				\toprule
				\multirow{2}{*}{Method} &  			
				\multicolumn{4}{c}{3D AP on Vehicle (IoU=0.7)}\\
				&Overall & 0 - 30 & 30 - 50 & 50 - inf\\
                \midrule
				2D Convolution &  63.57 & 84.64 & 59.54 & 38.58\\
        PointNet-RV~\cite{pointnet} &  63.47 & 84.43 & 59.32 & 38.29\\
        EdgeConv-RV~\cite{dgcnn} &  64.74 & 85.06 & 61.25 & 41.44\\
        ContinuousConv-RV~\cite{cc} & 63.52 & 84.47 & 59.63 & 38.40\\
        RSConv-RV~\cite{rsconv} &  63.47 & 84.45 & 59.70 & 38.13\\
        RandLA-RV~\cite{randla} &  64.11 & 84.95 & 60.17 & 39.06\\
        Meta-Kernel &  67.37 & 85.91 & 62.61 & 42.77\\
				\bottomrule
			\end{tabular}
		}	
	\end{center}
\vspace{-2mm}
  \caption{Comparison with point-based operators. The suffix ``RV'' means that the method is based on a fixed $3\times3$ neighborhood in RV instead of the dynamic neighborhood in 3D space. ContinuousConv in this table is the efficient version.}
	\label{tab:point_op}
\vspace{-2mm}
\end{table}

%% file: tables/diff_agg.tex
\begin{table}[H]
\small
\begin{center}
	\resizebox{0.9\linewidth}{!}{
		\begin{tabular}{l|c|c|c|c}
		\toprule
		$\mathcal{A}$ & Baseline & Max-pooling & Sum & Concate\\
		\midrule
		3D AP &63.57& 63.47 & 63.52 & 67.37\\
		\bottomrule
		\end{tabular}
	}
\end{center}
\vspace{-0.2cm}
\caption{Results of different aggregation strategies.}
\label{tab:diff_agg}
\vspace{-0.2cm}
\end{table}

%% file: rcp_ablation.tex
Instead of conditioning on the range, we try three other strategies to assign bounding boxes: azimuth span, projected area and visible area. The azimuth span of a bounding box is proportional to its width in the range image. The projected area is the area of a box projected into the range image. The visible area is the area of visible object parts. Note that area is the standard assign criterion in 2D detection. For a fair comparison, we keep the number of ground-truth boxes in a certain stride consistent between these strategies. Results are shown in Table \ref{tab:diff_condition}. We owe the inferior results to the pose change as well as occlusion, which makes the same object fall into different layers with different pose or occlusion conditions.
Such a result demonstrates that it is not enough to only consider the scale variation in the range image, since some other physical features, such as intensity, density, change with the range.
\input{tables/diff_condition.tex}

%% file: tables/diff_condition.tex
\begin{table}[H]
	\small 
	\begin{center}
		\resizebox{0.9\linewidth}{!}{
			\begin{tabular}{c|ccccc}
				\toprule
				\multirow{2}{*}{Conditions} &  			
				\multicolumn{4}{c}{3D AP on Vehicle (IoU=0.7)}\\
				&Overall & 0 - 30 & 30 - 50 & 50 - inf\\
                \midrule
				w/o RCP &  63.17 & 81.70 & 58.59 & 38.99\\
                Range &  67.37 & 85.91 & 62.61 & 42.77\\
                Span of azimuth &  64.04 & 80.63 & 62.28 & 42.34\\
                Projected area &  63.97 & 83.50 & 60.87 & 41.71\\
                Visible area &  59.43 & 79.69 & 57.69 & 34.67\\
				\bottomrule
			\end{tabular}
		}	
	\end{center}
	\vspace{-2mm}
  \caption{Comparison of different assignment strategies.}
	\label{tab:diff_condition}
	\vspace{-3mm}
\end{table}

%% file: wnms_ablation.tex
To support our claims in Sec. \ref{sec_wnms}, we apply weighted NMS in two typical voxel-based methods -- PointPillars~\cite{pointpillar} and SECOND~\cite{second} based on the strong baselines in MMDetection3D\footnote{\url{https://github.com/open-mmlab/mmdetection3d}}. Compared with RangeDet, the others generate fewer proposals due to memory and computational limits, which degrades the performance of weighted NMS as Table \ref{tab:wnms_on_diff_methods} shows.

%% file: ablation.tex
We further conduct ablation experiments on the components we use. Table \ref{tab:ablation} summarizes the results.
Meta-Kernel is effective and robust in different settings.
Both RCP and Weighted NMS significantly improve the performance of our whole system.
Although IoU prediction is a common practice of recent 3D detectors~\cite{pvrcnn,parta2}, it has a considerable effect on RangeDet, so we ablate it in Table \ref{tab:ablation}. 
\input{tables/wnms_on_diff_methods.tex}

%% file: tables/wnms_on_diff_methods.tex
\begin{table}[ht]
	\small 
	\begin{center}
		\resizebox{0.99\linewidth}{!}{
			\begin{tabular}{c|cccc}
				\toprule
				\multirow{2}{*}{Method} &  			
				\multicolumn{3}{c}{3D AP on Vehicle (IoU=0.7)}\\
				&RangeDet & \hspace{-2mm} PointPillars~\cite{pointpillar}\hspace{-2mm} & SECOND~\cite{second} \\
                \midrule
				NMS & 69.17 & 68.49 & 67.14 \\
                Weigted NMS  & 72.85 & 69.53 & 67.73 \\
                \midrule
				Improvements & +3.68 & +1.04 & +0.59 \\
				\bottomrule
			\end{tabular}
		}	
	\end{center}
	\caption{Results of weighted NMS on different detectors.}
	\label{tab:wnms_on_diff_methods}
	\vspace{-4mm}
\end{table}

%% file: sota.tex
Table \ref{tab:sota} shows that RangeDet outperforms other pure range-view-based methods, and is slightly behind the state-of-the-art BEV-based two-stage method.
Among all the results, we observe an interesting phenomenon: In contrast to the stereotype that range view is inferior in long-range detection, \emph{RangeDet outperforms most other compared methods in the long-range metric (i.e. 50m - inf), especially in the pedestrian class.} Unlike in the range view, the pedestrian is very tiny in BEV.
This again verifies the superiority of the range view representation and the effectiveness of our remedies to the inconsistency between range view input and 3D Cartesian output space.


%% file: runtime.tex
On Waymo Open Dataset, our model achieves 12 FPS evaluated on a single 2080Ti GPU without deliberate optimization.
Note that our method's runtime speed is not affected by the expansion of the valid detection distance, while the speed of BEV-based methods will quickly slow down as the maximum detection distance expands.